\documentclass[10pt,twocolumn,letterpaper]{article}

\usepackage[pagenumbers]{cvpr}      %

\usepackage{array}

\newcommand{\maybefig}[2]{\IfFileExists{#2}{\includegraphics[#1]{#2}}{%
  \fbox{\parbox{0.95\linewidth}{\centering\vspace{6mm}\textcolor{red}{Figure pending:
  \texttt{\detokenize{#2}}}\vspace{6mm}}}}}
\newcommand{\task}[1]{\textsc{#1}}

\definecolor{cvprblue}{rgb}{0.21,0.49,0.74}
\usepackage[pagebackref,breaklinks,colorlinks,allcolors=cvprblue]{hyperref}

\title{LabApprentice: Towards Harness-Level Recursive Self-Improvement for Wetlab Robotics}
\author{{\fontsize{11}{13.6}\selectfont Guoyue Xu$^{1,\dagger}$\hspace{0.8em}Mengzhao Jia$^{2,\dagger}$\hspace{0.8em}Yu Zhou$^{3,\dagger}$\hspace{0.8em}Qiang Jiang$^{1}$\hspace{0.8em}Yangyu Huang$^{1}$\hspace{0.8em}Renzheng Zhang$^{1}$}\\
{\fontsize{11}{13.6}\selectfont Bryan D. Paulsen$^{4}$\hspace{0.8em}Yanliang Zhang$^{1}$\hspace{0.8em}Brett Savoie$^{4}$\hspace{0.8em}Mengxue Hou$^{3,*}$\hspace{0.8em}Meng Jiang$^{2,*}$\hspace{0.8em}Tengfei Luo$^{1,4,*}$}\\[3pt]
{\normalsize $^{1}$Department of Aerospace and Mechanical Engineering \quad $^{2}$Department of Computer Science and Engineering}\\
{\normalsize $^{3}$Department of Electrical Engineering \quad $^{4}$Department of Chemical and Biomolecular Engineering}\\
{\normalsize University of Notre Dame}\\
{\normalsize $^{\dagger}$Equal contribution \quad $^{*}$Corresponding authors: {\tt mhou@nd.edu}, {\tt mjiang2@nd.edu}, {\tt tluo@nd.edu}}}

\begin{document}

\maketitle

\begin{abstract}
Wetlab robotics is a demanding setting for self-improving agents. Physical experiments take time and consume materials, leaving limited opportunities for trial and error. A wetlab measurement can tell the agent whether a procedure met its target, but not necessarily where it went wrong. We introduce LabApprentice, a wetlab robotics harness in which a coding agent recursively improves at the harness level. Without robot demonstrations, and without changing the model weights, the agent accumulates executable tools, written skills and bench measurements in a skill store that each run extends for the next. Our key insight is to use the laboratory's own instruments, unmodified, as step-level verifiers that confirm or refute a claim the agent records after each experimental step. On a Kinova Gen3 arm, we evaluate five tasks spanning common laboratory operations, including liquid handling, heating, vessel manipulation, and electrical measurement. We rerun each repeated task under a changed target, position or vessel orientation and compare against controls that start from an empty store. The agent measured the liquid that still arrived after a pour had stopped, stored a rule to stop that much earlier, and wrote a potentiostat program during execution. Agent-and-robot time fell by 22--59\% from the first to the last run of each repeated task, and single empty-store controls of pouring and splitting took 1.6--3.5$\times$ as long as the runs they repeated. The last task, a four-hour hydrogel workflow, combined the programs developed in the earlier tasks. The two resulting gels had conductivities of 0.71 and 1.12\,S/m, within the range reported for similar hydrogels, and a gel prepared by hand from the same recipe measured 0.71\,S/m. These results show that programs developed through separate physical experiments can be reused and composed to execute a complete wetlab workflow.
\end{abstract}

\section{Introduction}
\label{sec:intro}

Self-driving laboratories now run entire experimental campaigns~\cite{burger2020,szymanski2023,szymanski2026corr,abolhasani2023,tom2024}, but the procedures they execute still need to be programmed in advance. In the real lab, many experimental procedures share operations such as dispensing, heating, weighing and measuring, but their programs still need to be adapted to each new procedure's targets, vessels and bench layouts. Procedures requiring new instruments create additional integration work. For example, developing and debugging a mobile-robot workflow with unmodified laboratory instruments took more than a year~\cite{dai2024}. Chemical description languages such as $\chi$DL allow experimental procedures to run on different robotic platforms, but each platform needs to support the required operations~\cite{steiner2019,rauschen2024}. The challenge is therefore to turn this repeated programming effort into a library of reusable programs that an agent can build and revise through physical execution, using instrument readings to check the results.

\begin{figure*}[t]
  \centering
  \maybefig{width=\linewidth}{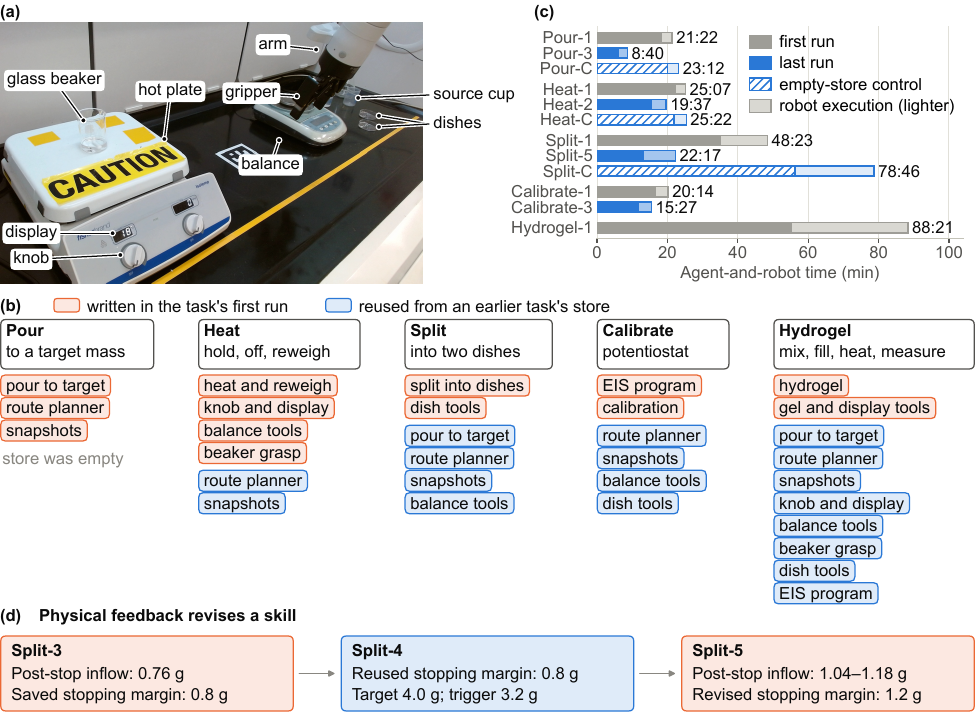}
  \caption{Overview. (a) The bench at the end of the fifth \task{Split} run, seen by a static camera. (b) The five tasks. Each task lists the skills and tools its first run stored (orange) and the items stored by earlier tasks' first runs that it ran (blue). Fig.~\ref{fig:provenance} lists every stored tool. EIS, electrochemical impedance spectroscopy. (c) Agent-and-robot time of each task's first run, last run with the store and empty-store control (Table~\ref{tab:runs}). The lighter end of each bar is robot execution and the rest is reasoning and programming. The operator stopped the \task{Calibrate} control, which is not drawn. (d) A stopping rule developed through \task{Split}. The third run measured liquid arriving after the stop and stored a 0.8\,g early-stop margin, the fourth reused it, and the fifth revised it to 1.2\,g.}
  \label{fig:overview}
\end{figure*}

Language models can generate robot programs from natural-language instructions~\cite{liang2023cap,singh2023progprompt}, and coding agents go further by writing, running and debugging programs over many steps, with a file system that can outlast a single task~\cite{agp2026,wetrobo2026}. The tools, files and instructions that surround the model form its harness. Instead of retraining a policy on its own data~\cite{bousmalis2023robocat}, such an agent can improve by keeping and editing the code it writes. Recursive self-improvement of this kind has mostly been shown in software, where an agent edits its own code and checks each change against benchmarks~\cite{stop2024,dgm2025,sica2025}, and recent work has begun to extend it to robots~\cite{embodirsi2026,embodiedevo2026,embodiedrsi2026}. Prior work has shown that stored programs and experience can speed up repeated tasks on real robots~\cite{agp2026,wetrobo2026,labevolver2026}.

 However, wetlab experiments remove the conditions that self-improvement in software relies on. Each attempt takes time and consumes materials, so there are no cheap resets or rollouts, and a final measurement may not reveal which step went wrong. We call this setting physical recursive self-improvement, in which verifying an improvement takes a physical measurement. Yet the laboratory's own instruments can provide feedback during execution. The question is whether an agent can use these readings to check and revise the programs it writes, and retain what it learns for future experiments.

We build on this idea with LabApprentice, a wetlab robotics harness in which a coding agent learns like an apprentice at a bench, working out a task on its first run, keeping what it wrote and getting faster on later runs. Because attempts are few, what a run learns should not be lost when its session ends. The agent's memory across runs is a skill store, akin to Voyager's skill library~\cite{voyager2023}, that holds tools (programs), skills (written instructions, in the sense of agent skills), notes and bench layouts. Self-improvement is at the harness level. The model weights stay fixed, and each run extends the store that the next run starts from, including tools that later runs use to build further tools. Because a failed outcome does not say which step failed, LabApprentice checks steps as they are carried out. Our key insight is to use the laboratory's instruments as step-level verifiers. After each experimental step, the agent records a claim, which the balance, the hot plate's display, the cameras or the potentiostat then confirm or refute, turning sparse outcome signals into process-level feedback. All instruments are used as delivered. The arm sets the hot plate by turning its knob while a camera reads the display, and the agent drives the potentiostat through the code it writes.

We evaluate LabApprentice on five tasks on a Kinova Gen3 arm (in Fig.~\ref{fig:overview}b), namely pouring to a target mass (\task{Pour}), heating and reweighing a beaker (\task{Heat}), splitting water between two dishes (\task{Split}), calibrating two conductivity cells with a potentiostat (\task{Calibrate}), and preparing and measuring a conductive poly(3,4-ethylenedioxythiophene) polystyrene sulfonate (PEDOT:PSS) hydrogel (\task{Hydrogel}). The operation types of these tasks account for 55--63\% of the operations annotated in three corpora of laboratory protocols~\cite{tamari2021xwlp,wang2022ulsa,vaucher2020actions}. Each later run changes the target, a position or the vessel's orientation, and each repeated task is rerun once from an empty store as an ablation, the empty-store control. Our cost metric is agent-and-robot time, reasoning and programming plus robot execution, reported with tokens and API cost.

In these runs, the agent converts measurements into stored rules. In \task{Split}, it measured liquid arriving after a stop, stored a rule to stop early by that amount, and in the next two runs hit the target in each dish with a single pulse (Fig.~\ref{fig:overview}d). With such rules and tools in the store, later runs are shorter. Agent-and-robot time fell by 22--59\% from the first to the last run of each repeated task, mostly in reasoning and programming, and single empty-store controls of \task{Pour} and \task{Split} took 1.6--3.5$\times$ as long. The stored tools also compose. The \task{Hydrogel} workflow ran 13 of the 14 stored tools, including the potentiostat program written in \task{Calibrate}, and produced two gels with conductivities of 0.71 and 1.12\,S/m, within the range reported for such hydrogels. Instrument readings also guide execution. In \task{Hydrogel}, camera frames showed the agent that a lid it believed it held lay on the bench, and it regrasped the lid and continued. In summary, we make three contributions.
\begin{itemize}\setlength{\itemsep}{0pt}\setlength{\parskip}{0pt}
  \item LabApprentice, a wetlab robotics harness for harness-level recursive self-improvement, in which a coding agent accumulates tools, skills and bench measurements for unmodified instruments, including a hot plate operated through its knob and display, with every arm motion gated by fixed safety checks.
  \item Instruments as step-level verifiers, through a claim-before-check protocol that logs a claim, a reading and a verdict for each experimental step.
  \item A controlled evaluation on five different tasks with bench changes between runs and an empty-store control for each repeated task.
\end{itemize}

\section{Related work}
\label{sec:related}

\paragraph{LLM agents in the laboratory.} Language-model agents already run chemistry on instruments with programming interfaces. Coscientist and LLM-RDF write instrument code from API documentation to drive liquid handlers~\cite{boiko2023,llmrdf2024}, ChemCrow planned and ran four syntheses on a cloud-connected platform~\cite{chemcrow2024}, AILA operates an atomic force microscope through its Python interface~\cite{aila2025}, and CRESt drives devices by emulating keyboard and mouse input to their software~\cite{crest2025}. ORGANA executes LLM-written plans with robot skills designed by experts~\cite{organa2024}. None of these systems retains the code it writes for later tasks.

\paragraph{Robots on instruments built for people.} An arm with a wrist camera can operate existing instruments through their physical controls~\cite{scirobotics2025,moran2026}. A dual-arm robot has run biology protocols with human pipettes and devices using motions programmed by experts~\cite{yachie2017}, and a robot has operated household appliances by reading their manuals and tracking their state from images~\cite{zhang2025apbot}. However, each step still needs a program. LLM agents that plan such work struggle to respect vessels, volumes and safety. When the workstations of an automated chemistry laboratory were exposed as agent-callable skills, experts judged 24.9\% of 607 dispatched protocols executable~\cite{ustcstress2026}.

\paragraph{Skill libraries and self-improving agents.} Self-improving coding agents edit their own code and validate each edit against software benchmarks~\cite{stop2024,dgm2025,sica2025}. Voyager stores each self-verified program in a skill library and, in a new Minecraft world, solved unseen tasks in 18--21 prompting iterations against 26--36 without it~\cite{voyager2023}. ASPIRE transfers repair skills from simulation to real-robot programming and learns text skills online on a real robot~\cite{aspire2026}, and Recursia refines shared tools, skill contracts and handoffs before reusing the frozen stack on new task compositions in simulation and on a Franka FR3~\cite{recursia2026}. On a Kinova Gen3, ELLMER retrieves curated robot code and raised plan faithfulness from 0.74 to 0.88, but its knowledge base is written by people and is not updated from experience~\cite{ellmer2025}. CaP-X synthesizes a skill library of nine functions from successful simulated rollouts in a single pass~\cite{capx2026}, and Guava distills an agent that calls the tools of a manipulation harness into a compact model fine-tuned on simulated trajectories~\cite{guava2026}. Three concurrent systems bring such self-improvement to robot manipulation. EmbodiRSI fine-tunes a vision-language-action policy round by round on simulated data that agents generate, and transfers it to a real dual-arm robot~\cite{embodirsi2026}. In EmbodiedEvo, whose experiments are all in simulation, a coding agent writes skills and lessons in one scene per task, and fresh agents given the frozen store succeed more often in unseen scenes than agents without it~\cite{embodiedevo2026}. EmbodiedRSI evolves the code and skills of an agentic harness in simulation and runs the frozen harness on a real arm~\cite{embodiedrsi2026}. In all three, unlike ENPIRE~\cite{enpire2026}, the improvement loop runs in simulation. None of these systems operates laboratory instruments.

\paragraph{Experience reuse on real robots.} In AGP, each execution starts from the measurements, programs and lessons saved from earlier attempts, and repeated runs became faster mainly through less reasoning and programming. However, its tasks lay outside the laboratory, success in the reuse runs rested on the agent's self-report or an operator's review, and its authors note that isolating the speed-up requires repeats from empty experience~\cite{agp2026}. ARCHITECT distills a person's natural-language corrections to agent-written programs into a persistent skill library on a Franka arm, and with the library populated, human queries fell from 4.67 to 0.83 per trial~\cite{architect2026}. LEXP retains lessons from real-robot executions with the model weights fixed, and on three tasks removing the retained lessons reduced success from 7 of 9 trials to 3~\cite{lexp2026}. Neither involves laboratory instruments. WetRobo reopened an incubator door in 3.5\,min after a first opening of 2\,h\,40\,min, but left open whether its programs help when the setting changes~\cite{wetrobo2026}. LabEvolver tunes the parameters of skills designed by people, including the pre-stop of a pouring controller, and halved the reagent additions of a pH and conductivity adjustment with stored experience~\cite{labevolver2026}. In LabApprentice, the agent keeps the code it writes, revises it on the real bench and operates laboratory instruments as delivered. The evaluation combines empty-store controls, bench changes between runs and instrument verification of each step (Supplementary Table~S5).

\section{LabApprentice}
\label{sec:method}

\begin{figure*}[t]
  \centering
  \maybefig{width=\linewidth}{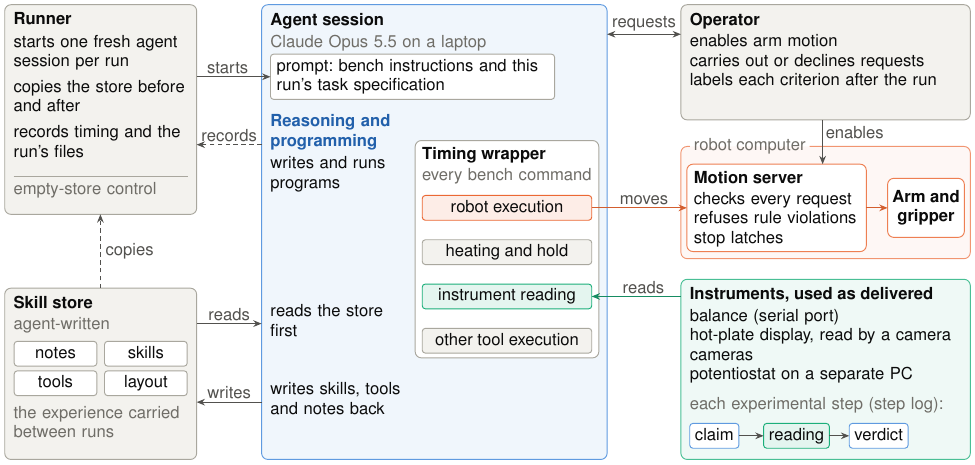}
  \caption{Structure of LabApprentice. The runner starts one fresh agent session per run with the bench instructions and that run's task specification, and copies the agent-written skill store before and after the run. The agent reads the store first, writes and runs its programs, sends every bench command through a timing wrapper, moves the arm through the motion server, and logs a claim, an instrument reading and a verdict for each experimental step. The operator at the bench answers logged requests and labels the criteria after the run.}
  \label{fig:harness}
\end{figure*}

\paragraph{Setting.} Run $k$ executes one task variant in a fresh agent session whose memory is the skill store $\mathcal{S}_k$. The run produces a trajectory $\tau_k$ of programs, bench commands and instrument readings, and ends with the agent writing $\mathcal{S}_{k+1} = U(\mathcal{S}_k, \tau_k)$, where the agent performs the update $U$. The model weights $\theta$ stay fixed. We call this harness-level self-improvement, and recursive because $\mathcal{S}_k$ contains the tools with which run $k$ builds $\mathcal{S}_{k+1}$. An empty-store control runs the same variant from $\mathcal{S} = \emptyset$.

\paragraph{Harness and skill store.} We provide the runner, the motion server, the timing wrapper, three skills written before the first run and the bench instructions, which serve as the system prompt (Fig.~\ref{fig:harness}). These programs and skills stayed fixed within each task. The store, written by the agent, holds a notes file, skills (written instructions for a kind of task), tools (programs, each documented with its interface and what it was tested on) and layout files (dated bench positions and calibrations). In each run, the agent receives a task specification with the run's parameters, reads the store, writes the programs it lacks, such as stopping a pour at a target mass, carrying a Petri dish onto the balance pan or starting an impedance measurement, and before finishing stores what the next run should reuse. The runner copies the store before and after every run, and each empty-store control writes into a fresh store.

\paragraph{Session isolation.} Each run was a new session running Claude Opus~5.5 at high reasoning effort, without saved sessions and with user instructions, skills, hooks and memory disabled. The bench instructions put task specifications, other runs' records and stores and the experimenters' files off limits. A script checked every command against these rules, and no run read another run's records or data.

\paragraph{Instruments as step-level verifiers.} Errors compound along a workflow. At 99\% success per operation, a workflow of 20 operations succeeds about 82\% of the time~\cite{scirobotics2025}. LabApprentice therefore verifies each experimental step as well as the final outcome. The task specification names an instrument for each step. Under a claim-before-check protocol, the agent logs a claim about what it believes happened before reading that instrument, and then logs the reading and its verdict. Instruments, criteria and labels lie outside the agent's reach, and after each run the operator labels every criterion (Supplementary Table~S3) from the instrument records and the agent's checks.

\paragraph{Perception.} The agent sees the bench through still frames from a wrist camera and two fixed cameras, which it opens as images. To locate a vessel, it reads pixel coordinates off a gridded crop of a wrist frame, and stored tools convert them to bench coordinates with the camera model and the arm's pose. No run with the store used fiducial markers or depth images. The agent judges a grasp from frames taken before and after a short test lift, and it reads the hot plate's display from camera frames.

\paragraph{Motion server.} Because the agent writes its motion code at run time, every arm motion passes through a single ROS~2 process with fixed safety checks. The server accepts requests to stop, to operate the gripper, or to move the flange along a Cartesian straight line or the arm along a joint-space line, in which all joints move at proportional rates. It executes motions while an enable gate, set by the operator when the server starts, is open, and refuses them otherwise. It rejects a motion issued during another motion, starting more than 2$^\circ$ from the measured joint state, outside joint limits or faster than 30$^\circ$/s. Forward kinematics checks every joint-space move, and Cartesian moves given limits, to keep the tool center point above a floor plane and outside the collision boxes that the agent sends with the request. A stop is never rejected.

\section{Experiments}
\label{sec:experiments}

The five tasks show what the agent stores after a run and how later runs change it (Section~\ref{sec:r-skills}), and how much the store shortens later runs (Section~\ref{sec:r-repeat}). The runs also test whether tools stored by separate tasks can be combined into a new experiment (Section~\ref{sec:r-transfer}) and whether instrument readings guide execution (Section~\ref{sec:r-checks}).

\subsection{Setup}
\label{sec:setup}
\paragraph{Hardware.} The bench is arranged around a Kinova Gen3 arm with a Robotiq 2F-85 parallel gripper, a wrist camera and two fixed cameras, a RealSense D435 and a Logitech C920 (Fig.~\ref{fig:overview}a). The balance streams its reading over a serial line and accepts no read request, the stirring hot plate has knobs and seven-segment displays but no computer interface, and the potentiostat is reached through the vendor's Windows toolkit. Every instrument was used as delivered. The vessels were a 50\,mL polypropylene cup, a 100\,mL glass beaker with a stir bar and plastic Petri dishes about 53\,mm across with loose lids. Two dishes, the molds, served as two-electrode conductivity cells, each with two flat copper-foil electrodes on its bottom (Supplementary Fig.~S2). One pair served in \task{Calibrate} and a new pair in \task{Hydrogel}.

\begin{table*}[tp]
  \centering
  \caption{The five tasks. Parameters of each numbered variant, the change on the bench before its runs, measured from the run records (Supplementary Section~S1), and its runs. C marks the empty-store control.}
  \label{tab:ladder}
  \footnotesize
  \begin{tabular}{@{}>{\raggedright\arraybackslash}p{0.10\textwidth}>{\raggedright\arraybackslash}p{0.04\textwidth}>{\raggedright\arraybackslash}p{0.27\textwidth}>{\raggedright\arraybackslash}p{0.30\textwidth}>{\raggedright\arraybackslash}p{0.18\textwidth}@{}}
    \toprule
    Task & Var. & Parameters & Change on the bench & Runs \\
    \midrule
    \task{Pour} & 1 & 20.0\,$\pm$\,1.0\,g & & \task{Pour}-1 \\
     & 2 & 30.0\,$\pm$\,1.0\,g & cup moved & \task{Pour}-2, \task{Pour}-C \\
     & 3 & 25.0\,$\pm$\,1.0\,g & cup moved & \task{Pour}-3 \\
    \midrule
    \task{Heat} & 1 & 45\,$^\circ$C, hold 5\,min & & \task{Heat}-1 \\
     & 2 & 40\,$^\circ$C, hold 5\,min & hot plate moved  & \task{Heat}-2, \task{Heat}-C \\
    \midrule
    \task{Split} & 1 & 4.0\,$\pm$\,0.5\,g per dish &  & \task{Split}-1, \task{Split}-4 \\
     & 2 & 3.0\,$\pm$\,0.5\,g per dish & dishes moved & \task{Split}-2, \task{Split}-5, \task{Split}-C \\
     & 3 & 4.0\,$\pm$\,0.5\,g per dish & dishes moved & \task{Split}-3 \\
    \midrule
    \task{Calibrate} & 1 & 8.0\,g of standard per mold &  & \task{Calibrate}-1 \\
     & 2 & 7.0\,g of standard per mold & molds moved & \task{Calibrate}-2, \task{Calibrate}-C \\
     & 3 & 9.0\,g of standard per mold & molds moved & \task{Calibrate}-3 \\
    \midrule
    \task{Hydrogel} & 1 & 16.0\,$\pm$\,0.5\,g PEDOT:PSS, 800\,rpm for 2\,min, 6.0\,$\pm$\,1.0\,g per mold, 60\,$^\circ$C for 2\,h & hot plate moved & \task{Hydrogel}-1 \\
    \bottomrule
  \end{tabular}
\end{table*}

\paragraph{Tasks.} The last of the five tasks is a hydrogel workflow that combines the earlier four (Table~\ref{tab:ladder}). The five tasks exercise operation types, such as transfer, heating, mixing and measurement, that account for 63\% of the operations annotated in 279 wetlab biochemistry protocols~\cite{tamari2021xwlp}, 61\% of the actions in 535 inorganic synthesis paragraphs~\cite{wang2022ulsa} and about 55\% of those in 352 sentences of organic synthesis procedures~\cite{vaucher2020actions} (Supplementary Table~S7). Each task specification states the goal, starting state, parameters, criteria with their instruments, constraints and common rules (write a missing capability into the store, edit no harness file outside the store, record failures as failures), and the prompt carries one numbered variant. The agent was also instructed to ask the operator for anything outside the arm's reach or unexpected, such as refills, vessel exchanges, cable changes at the potentiostat and checks of the bench, and the operator carried out or declined each request. In \task{Calibrate}, the agent writes a program that runs an impedance measurement and reads the result, and determines each mold's cell constant $K$, which converts its resistance $R$ into a conductivity, from a 1000\,$\pm$\,10\,$\mu$S/cm sodium chloride standard that the operator fills into the molds. Before \task{Hydrogel}, we calibrated the new molds with stirred standard (Supplementary Fig.~S3).

\paragraph{Protocol.} Within a task, runs differ in the store, the target and the bench layout (Table~\ref{tab:ladder}), and each is a new session with the same model and effort. Runs are named by task and order and took place in the order of Table~\ref{tab:runs}. An empty-store control reruns a task's second variant from an empty store, after the task's runs with the store, with the same task specification, operator and bench.

\begin{table}[t]
  \centering
  \caption{Runs of the five tasks. Columns give agent-and-robot time (Agent+robot, reasoning and programming plus robot execution, in min:s), the model's thinking tokens (Think), API cost (Cost) and the success criteria labeled from the instrument records (Passed). Runs ending in C are empty-store controls, and all others used the skill store. The operator stopped the \task{Calibrate} control. \task{Hydrogel} is timed to the end of its first round of measurements, and its tokens and cost cover the whole session.}
  \label{tab:runs}
  \footnotesize
  \setlength{\tabcolsep}{3pt}
  \begin{tabular}{@{}lccrrr@{}}
\toprule
Run & Var. & Passed & Agent+robot & Think (k) & Cost (\$) \\
\midrule
Pour-1 & 1 & 2/2 & 21:22 & 45.6 & 8.79 \\
Pour-2 & 2 & 2/2 & 10:15 & 15.4 & 2.68 \\
Pour-3 & 3 & 2/2 & 8:40 & 10.1 & 2.30 \\
Pour-C & 2 & 2/2 & 23:12 & 35.1 & 8.52 \\
Heat-1 & 1 & 4/4 & 25:07 & 43.8 & 9.81 \\
Heat-2 & 2 & 4/4 & 19:37 & 27.2 & 7.64 \\
Heat-C & 2 & 4/4 & 25:22 & 45.4 & 10.98 \\
Split-1 & 1 & 2/3 & 48:23 & 105.7 & 16.95 \\
Split-2 & 2 & 2/3 & 48:12 & 100.5 & 14.11 \\
Split-3 & 3 & 2/3 & 32:23 & 58.5 & 9.06 \\
Split-4 & 1 & 3/3 & 26:44 & 37.3 & 7.21 \\
Split-5 & 2 & 3/3 & 22:17 & 24.3 & 5.59 \\
Split-C & 2 & 1/3 & 78:46 & 171.9 & 31.00 \\
Calibrate-1 & 1 & 2/2 & 20:14 & 34.4 & 10.43 \\
Calibrate-2 & 2 & 2/2 & 14:44 & 18.4 & 4.58 \\
Calibrate-3 & 3 & 2/2 & 15:27 & 19.5 & 4.74 \\
Calibrate-C & 2 & 1/2 & 34:06 & 92.4 & 14.30 \\
Hydrogel-1 & 1 & 8/8 & 88:21 & 145.6 & 36.21 \\
\bottomrule
\end{tabular}

\end{table}

\paragraph{Metrics.} A timing wrapper around every bench command records its span and category, as robot execution, instrument reading (including the potentiostat), procedure wait (heating up, the hold, stirring) or other tool execution. The remaining time, excluding waits for the operator, is reasoning and programming, in which the agent reads, thinks, writes code and waits for the model. Because the task specification fixes the procedure waits, we compare runs on agent-and-robot time, the sum of reasoning and programming and robot execution, over the whole session. \task{Hydrogel} is timed to the end of its first round of measurements. Supplementary Table~S2 reports total elapsed time, operator requests and interventions. Task success is the number of criteria met. The 19 criteria, with their instruments and operational definitions, are in Supplementary Table~S3 and cover poured mass and no spill (\task{Pour}), setpoint, hold, OFF and mass on reweighing (\task{Heat}), mass in each dish and no spill (\task{Split}), the measurement program and cell constants (\task{Calibrate}), and eight checks in \task{Hydrogel}, from the added mass to a conductivity for each mold. Supplementary Section~S4 gives full details.

\subsection{From bench measurements to stored tools}
\label{sec:r-skills}
\begin{figure*}[t]
  \centering
  \maybefig{width=\linewidth}{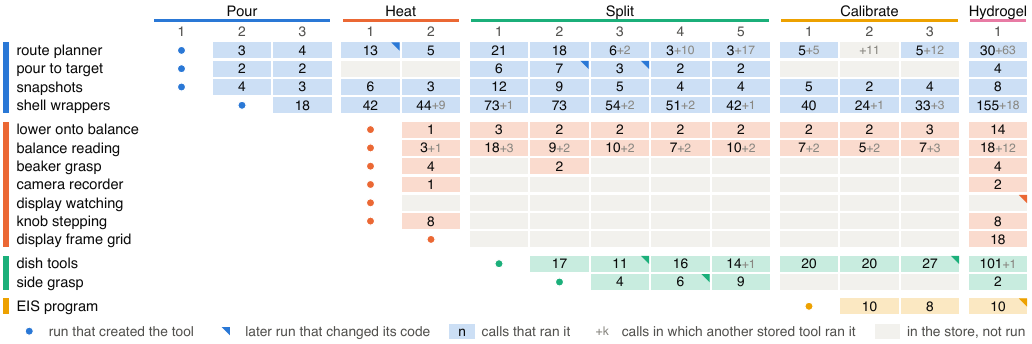}
  \caption{Which run ran which stored tool. Rows are the tools in the store, grouped and colored by the task whose run wrote them, and columns are the runs with the store in run order. A dot marks the run that created a tool, a corner triangle a later run that added to or changed its code, and a number the calls that ran the tool. A gray +k counts calls in which the tool ran because another stored tool called or imported it, and gray cells are tools that were in the store but not run.}
  \label{fig:provenance}
\end{figure*}

\begin{figure}[t]
  \centering
  \maybefig{width=\linewidth}{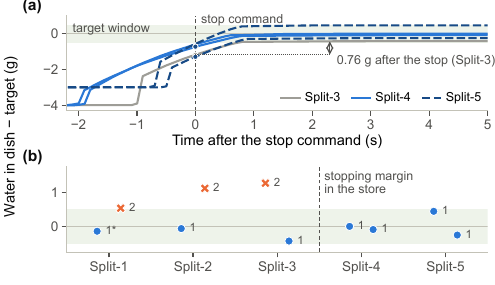}
  \caption{Physical feedback revises a stored rule in \task{Split}. (a) Balance readings during five pulses of the third, fourth and fifth runs, relative to each dish's target, with time measured from the stop command. (b) Water in both dishes of every run relative to the target, with the number of pulses beside each point. Crosses mark dishes outside the 0.5\,g window, and the asterisk a dish filled after the operator had emptied two overshoots.}
  \label{fig:stoprule}
\end{figure}

The agent revised stored tools and skills in response to what it measured, and later runs extended or changed six of the 14 stored tools (Fig.~\ref{fig:provenance}). In \task{Split}, the agent extended the pour-to-target tool inherited from \task{Pour}. In the third run, it measured about 0.8\,g arriving after a stop and stored a rule to stop that much before the target. It applied the rule to both dishes in the fourth run (Figs.~\ref{fig:overview}d and~\ref{fig:stoprule}a). With the rule, the agent filled each dish in a single pulse in the fourth and fifth runs and met both mass windows, whereas none of the first three runs met both (Fig.~\ref{fig:stoprule}b).

As experimental targets, vessels, and bench layouts changed, the agent reused and extended its stored tools when new operations were needed. The cup in \task{Pour}, the hot plate in \task{Heat} and the dishes in \task{Split} moved between variants (Table~\ref{tab:ladder}). The second \task{Heat} run reused its stored grasp pose and reached the moved plate with checked joint-space moves. When the beaker's spout faced the other way in the second \task{Split} run, the agent wrote a tool for the mirrored side grasp, and in every later run with the store each grasp of the beaker held at the first close.

The agent also brought a new instrument into use. Before \task{Calibrate}-1, the harness contained no potentiostat code. The agent read the vendor's help files and example script, wrote a program that runs an electrochemical impedance spectroscopy (EIS) sweep and reads its result, and completed its first sweep 14.5\,min into the run, excluding waits for the operator. It stored the program and a mold-calibration skill, which the second and third runs reused to complete their first sweeps after 6.4 and 8.0\,min. The program gave repeatable readings. In each \task{Calibrate} run with the store, the two readings of each mold differed by less than 0.5\%.

\subsection{Effect of the skill store}
\label{sec:r-repeat}
In all four repeated tasks, the last run needed less agent-and-robot time than the first (Fig.~\ref{fig:overview}c, Table~\ref{tab:runs}). The reduction was largest in \task{Pour}, 59\% from 21:22 to 8:40, and smallest in \task{Heat}, 22\%, with its hot plate moved farther from the arm. In \task{Split}, whose fourth and fifth runs repeated the targets of the first two, time fell by 45\% at 4.0\,g per dish and by 54\% at 3.0\,g, and the second and third \task{Calibrate} runs were 27 and 24\% shorter than the first. From the first run to the last, thinking tokens fell by 38--78\% and API cost by 22--74\%. As the runs shortened, every run of \task{Pour} and \task{Heat} still met all its criteria (Supplementary Table~S3), and operator requests fell from three in each of the first two \task{Split} runs to none in the fifth.

\paragraph{Where the saving comes from.} Breaking down each run's time (Fig.~\ref{fig:overview}c and Supplementary Fig.~S4) shows that reasoning and programming fell from 18:17 to 6:06 in \task{Pour} and from 34:59 to 13:03 in \task{Split}, and accounted for 84\% or more of the saving in each of the four tasks. As programming shrank, robot motion became a larger share of each run, reaching 2:34 of the 8:40 of \task{Pour}-3 and 9:14 of the 22:17 of \task{Split}-5.

\paragraph{Empty-store controls.} To test whether the store accounts for this saving, we reran one variant of each repeated task from an empty store. None of the three empty-store controls that ran to completion was as fast as the run with the store that it repeated, suggesting that the saving comes from the store. At 30\,g, the \task{Pour} control took 23:12 against 10:15 for \task{Pour}-2 (2.3$\times$), and both met their two criteria. At 3\,g per dish, the \task{Split} control took 78:46 against 48:12 for \task{Split}-2 and 22:17 for \task{Split}-5 (1.6$\times$ and 3.5$\times$), and met one of three criteria against two and three. \task{Split}-2 started without the mirrored grasp and the early-stop rule and \task{Split}-5 with both, suggesting that these additions account for part of the gap between them. The \task{Heat} control took 25:22 against 19:37 for \task{Heat}-2. The operator stopped the \task{Calibrate} control after it cracked a mold, before either mold had been measured, and we exclude its 34:06 from these comparisons.

With the store, repeated runs took 22--59\% less agent-and-robot time, mostly through less reasoning and programming, and empty-store controls took 1.3--3.5$\times$ as long.

\subsection{Composition into a long-horizon workflow}
\label{sec:r-transfer}
Each task started from the tools that earlier tasks had stored, and the execution records show which it ran (Fig.~\ref{fig:provenance}). In \task{Split}-1, 96\% of bench commands ran a stored tool, including the pour-to-target tool from \task{Pour} and the balance tools from \task{Heat}. Stored tools were also built from earlier ones, which is the recursion defined in Section~\ref{sec:method}. For example, the dish tool written in \task{Split} imports code from the route planner written in \task{Pour}.

\task{Hydrogel} ran 13 of the 14 stored tools. The pouring tool (\task{Pour}) supplied liquid handling, the knob and display tools (\task{Heat}) stirring and heating, the dish tools (\task{Split}) mold handling and the EIS program (\task{Calibrate}) electrical measurement. With these tools, the agent mixed PEDOT:PSS, an aqueous dispersion of a conductive polymer, into the precursor stock and filled two molds on the balance. The agent then heated the covered molds for 2\,h at a 60\,$^\circ$C display target and measured each gel with the potentiostat. Excluding waits for the operator, the run took 3\,h\,39\,min, of which 1\,h\,28\,min was agent-and-robot time, 2\,h\,1\,min procedure waits and the remaining 10\,min instrument reading and other tool execution. All eight criteria were met (Supplementary Table~S3), with 15.72\,g of PEDOT:PSS added against a target of 16.0\,g and 5.26 and 5.62\,g poured into the molds.

The two gels had resistances of 142.3 and 115.8\,$\Omega$, which with the molds' cell constants give conductivities of 0.71 and 1.12\,S/m (Supplementary Section~S4). For comparison, we prepared a gel by hand from the same recipe and cured it in the second mold. With the same EIS program this gel gave 0.71\,S/m, equal to the lower of the agent's two values. All three values lie within the 0.01--2.2\,S/m reported for PEDOT:PSS hydrogels blended with hydrogel precursors~\cite{feig2018}, within the 0.2--2.2\,S/m of other blends~\cite{wu2017}, and close to the 0.65\,S/m of a blended gel cured at 60\,$^\circ$C~\cite{pan2026}.

\subsection{Instrument verification}
\label{sec:r-checks}
Instrument readings guided execution as well as verifying it. The pour tool used the balance stream, including the stored early-stop margin, to decide when to stop, and the agent read the display between knob turns to decide whether the setpoint or OFF had been reached. For the two-hour hold in \task{Hydrogel}, the agent wrote a template classifier for the display, labeled its reference crops itself, classified 7,388 frames and inspected the 216 it flagged. Across the store, seven of the 14 tools handle camera frames or pixel coordinates, and the model viewed between 10 and 145 camera images in a run. Camera frames also let the agent recover from an error in \task{Hydrogel}. When a lid it believed it held was missing from its mold, it found the lid on the bench in its camera frames, regrasped it and confirmed the grip with a test lift before carrying it, and it placed the other lid the same way at the first attempt.

\section{Discussion}
\label{sec:discussion}

In five tasks on one bench, LabApprentice shows that harness-level recursive self-improvement carries over from software to the physical setting of wetlab robotics. Without weight updates, programs and operating knowledge acquired during execution became tools that later tasks reused and extended. The agent revised its grasp and stopping rules in \task{Split} as vessel orientations and pouring behavior changed, later \task{Pour} and \task{Split} runs executed stored programs where their empty-store controls rewrote them, and tools written in \task{Pour}, \task{Heat}, \task{Split} and \task{Calibrate} together served the hydrogel workflow. Because the experience is stored as code and notes, a person can inspect and correct what the robot has learned.

For laboratories, such reuse makes automation cumulative and changes when its cost is paid. Instead of experts programming each procedure up front~\cite{dai2024}, the first run of a task carries most of the programming, later runs reuse it, and a new task starts from what earlier tasks stored. It also gives self-driving laboratories a path to instruments built for people. Planning agents such as Coscientist and CRESt choose experiments but reach devices through APIs or their software~\cite{boiko2023,crest2025}. For instruments that offer neither, a skill store acquired on the bench could serve such a planner as its execution layer.

\paragraph{Limitations.} Each condition was run once, so each control ratio rests on a single pair of runs, and the controls ran after the runs with the store. The study used one bench, one operator and one model. Transferring stores across benches and laboratories and adding new classes of instruments are natural next steps.

\section*{LLM Usage Statement}
The system studied in this paper is driven by an LLM. Claude Opus 5.5 (Anthropic) wrote and ran the robot and instrument programs during the reported runs, as Sections 3 and 4 describe. Coding agents built on Claude models (Anthropic) wrote and revised the programs that we provide and the analysis and plotting scripts under our direction, and we reviewed and tested the code. Coding agents also helped to prepare and monitor the runs and computed the reported times, counts and tables from the run records. We designed the experiments, operated the bench and labeled every criterion. The plots and diagrams were produced by scripts from our designs. In writing the paper, we used the same tools to propose and revise text in every section and to check it against the run records. No LLM was used to create synthetic data or to label the criteria. In one run the agent labeled the reference crops of its own display classifier (Section~\ref{sec:r-checks}). The authors take full responsibility for all content.
{
    \small
    \bibliographystyle{ieeenat_fullname}
    \bibliography{references}
}

\end{document}


\maketitle

\section{What changed between runs: positions, spout direction and grasp}
\label{si:changes}

\begin{figure}[!htb]
  \centering
  \maybefig{width=\textwidth}{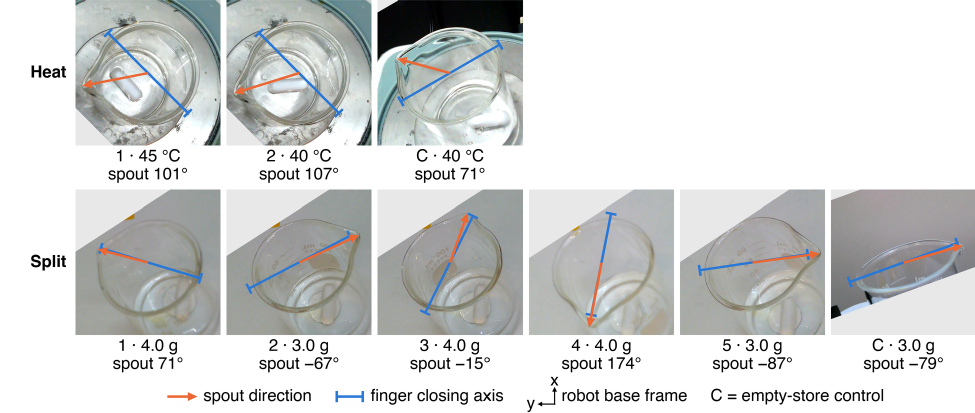}
  \caption{The source vessel in the wrist-camera frame that each run of \task{Heat} and \task{Split} took before grasping. Each task has one row, with the runs in order (C marks the empty-store control). Every view is rotated so that the robot base $+x$ points up and $+y$ points left, and light gray marks the area outside the camera's view. The arrow gives the spout direction, and the line with end bars gives the axis along which the fingers close. The panel of the \task{Split} control is a side view taken at the grasp pose. In \task{Heat}, the second run reused the first run's grasp pose, so the finger closing axis is the same in both panels. The control found a different grasp. In \task{Split}, the spout faced a different way in every run, and the finger closing axis turned with it.}
  \label{fig:si-frames}
\end{figure}

This section shows what changed on the bench between runs. For every run of \task{Heat} and \task{Split}, Fig.~\ref{fig:si-frames} shows the wrist-camera frame taken before the grasp, with the azimuth of the spout under each frame. For every run of \task{Pour}, \task{Heat} and \task{Split}, Table~\ref{tab:si-grasp} gives the orientation of the grasp. Azimuths are measured in the robot base frame, counter-clockwise from $+x$.

\paragraph{\task{Pour}.} The target changed in every run, and the operator moved the source cup before the third run. Its spout faced the balance in all four runs, at azimuths of 76$^\circ$ to 85$^\circ$. In each run the gripper used the same grasp from the side, with the spout above one finger, and its approach direction followed the spout.

\paragraph{\task{Heat}.} The operator moved the hot plate before the second run. The second run reused the grasp pose stored by the first run and reached the moved plate with joint-space moves that the motion server checked before the arm moved. In both runs the arm stopped turning the knob when the display showed the setpoint. The control found a different grasp, with the tool tilted 40$^\circ$.

\paragraph{\task{Split}.} The dishes were moved for each variant, and the beaker's spout faced a different way in every run. In every run the gripper held the beaker with the spout above one finger, so its approach direction followed the spout. The second run wrote a tool that computes the mirrored side grasp for any spout direction. In every later run with the store, each grasp of the beaker held at the first close.

\paragraph{\task{Hydrogel}.} The hot plate's center moved from the position recorded in the store, near the edge of the arm's reach, so both molds had to go on the half of the plate nearer the robot. The agent therefore stirred the mixture on the plate, returned the beaker to the bench for the pours and kept the plate free for the molds. The one grasp that could reach mold~1 passed close to the cup of PEDOT:PSS, so the arm first moved the cup 5\,cm toward the robot base. The agent then carried mold~2 to the balance through the place where mold~1 had stood.

\begin{table}[!htb]
  \centering
  \caption{Grasp in every run of \task{Pour}, \task{Heat} and \task{Split}. For the side grasps of \task{Pour} and \task{Split}, Grasp gives the azimuth from which the gripper approached and the finger under the spout. For the top grasps of \task{Heat}, Grasp gives the wrist yaw $\psi$ and the tool tilt. Closes counts how many times the gripper closed on the source vessel in the run.}
  \label{tab:si-grasp}
  \small
  \begin{tabular}{@{}llll@{}}
    \toprule
    Task & Run & Grasp & Closes \\
    \midrule
    \task{Pour} & 1 & approach $-$9$^\circ$, spout over $+x$ finger & 1 \\
       & 2 & approach $-$14$^\circ$, same & 1 \\
       & 3 & approach $-$6$^\circ$, same & 1 \\
       & C & approach $-$5$^\circ$, same, other wrist configuration & 1 \\
    \midrule
    \task{Heat} & 1 & top grasp, $\psi$ $-$45$^\circ$, tilt 21$^\circ$ & 1 \\
       & 2 & same pose, reused from run 1 & 1 \\
       & C & top grasp, $\psi$ $+$30$^\circ$, tilt 40$^\circ$ & 1 \\
    \midrule
    \task{Split} & 1 & approach $-$21$^\circ$, $+x$ finger under the spout & 5 \\
       & 2 & approach 27$^\circ$, $-x$ finger & 3 \\
       & 3 & approach 68$^\circ$, $-x$ finger & 2 \\
       & 4 & approach 84$^\circ$, $+x$ finger & 2 \\
       & 5 & approach 4$^\circ$, $-x$ finger & 2 \\
       & C & approach 20$^\circ$, $-x$ finger & 6 \\
    \bottomrule
  \end{tabular}
\end{table}

\section{The hydrogel molds}
\label{si:molds}

\begin{figure}[!htb]
  \centering
  \maybefig{width=\textwidth}{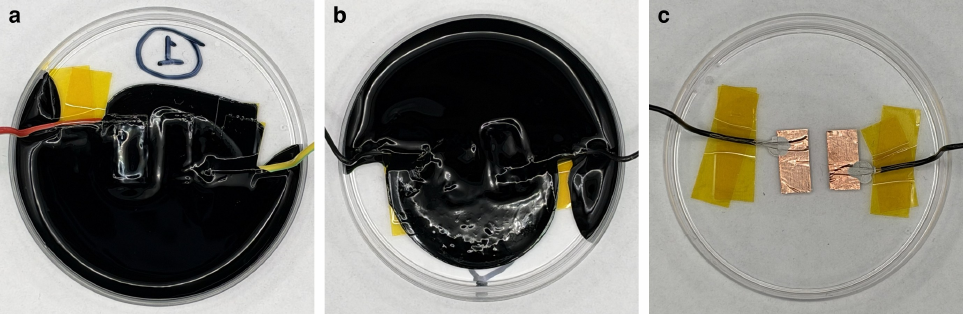}
  \caption{The hydrogel molds. (a)~Mold~1 and (b)~mold~2 a few minutes after the \task{Hydrogel} run ended, with the lids off. (c)~An empty mold of the same design, with its two flat electrodes on the bottom of the dish.}
  \label{fig:si-molds}
\end{figure}

\begin{figure}[!htb]
  \centering
  \maybefig{width=\textwidth}{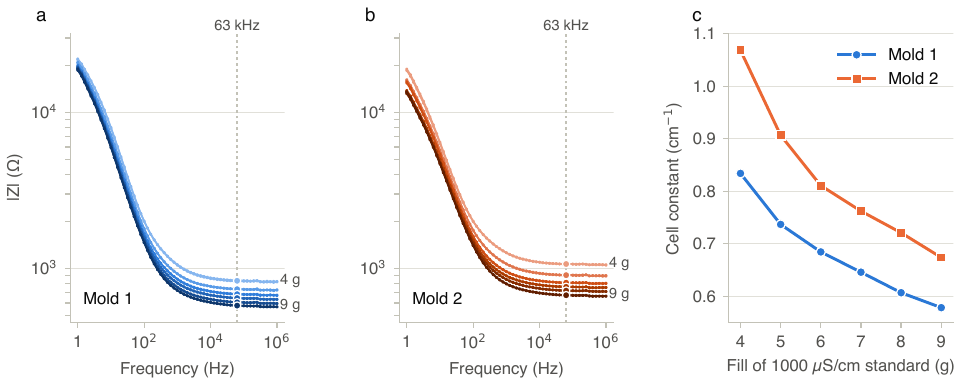}
  \caption{Calibration of the molds used for the gels with stirred 1000\,$\mu$S/cm sodium chloride standard on the balance pan. (a,~b)~Impedance magnitude $|Z|$ from 1\,MHz to 1\,Hz, with two readings per fill, light to dark from 4 to 9\,g. Dots mark the solution resistance $R$, the real part at 63\,kHz (dashed line). The 79\,kHz point, an instrument artifact, is not drawn. (c)~Cell constant $K = R\,\kappa_{\mathrm{std}}$, so that $K$ in cm$^{-1}$ equals $R$ in k$\Omega$.}
  \label{fig:si-cells}
\end{figure}

Each mold is a Petri dish about 53\,mm across with two flat electrodes on its bottom, wired to the potentiostat as a two-electrode cell. The leads run over the rim and are held by tape, and mold~1 has red and yellow leads, mold~2 black ones (Fig.~\ref{fig:si-molds}). Figure~\ref{fig:si-cells} shows the calibration of the two molds used for the gels, which Section~\ref{si:m-hydrogel} describes.

\section{Help from the operator}
\label{si:help}

Table~\ref{tab:si-help} lists total elapsed time, time waiting for the operator, requests grouped by reason and the operator's interventions over the timed period of each run. Elapsed time includes agent work, robot and instrument execution, procedure waits and waiting for the operator. A cable request asks the operator to connect, disconnect or move the potentiostat's cell cable, which the task specification assigns to the operator. A clean-up request asks the operator to pour back, discard or wipe water or to refill a vessel, and a vessel request asks the operator to place a vessel. A permission request asks whether a step outside the bench instructions is allowed. A fault request asks the operator to clear a stop or a fault, and a check request asks the operator to look at the bench or at a measurement. Cable changes at the potentiostat and refills still needed the operator. An intervention is work the operator did on the bench or its computers beyond answering a request. Waiting covers the time until the operator answered and the time that the operator spent helping, and the operator's hands-on work was not timed separately.

For \task{Hydrogel}, the timed period ends after the first round of measurements. It lasted 4\,h\,00\,min\,48\,s, of which the agent waited 21\,min\,25\,s for the operator. The whole session, with the later measurements, lasted 5\,h\,26\,min\,11\,s, and the thinking tokens and API cost cover the whole session. For \task{Calibrate}-C, the timed period ends when the operator stopped the failed run.

\begin{table}[!htb]
  \centering
  \caption{Elapsed time and help from the operator in every run. Elapsed is the total duration of the timed period, and Waiting is the time waiting for the operator, which Elapsed includes. Both are in min:s. Interv.\ counts the operator's interventions. \task{Hydrogel} ends at its first measurement round, and \task{Calibrate}-C at the stop.}
  \label{tab:si-help}
  \footnotesize
  \setlength{\tabcolsep}{3pt}
  \begin{tabular}{@{}lrr>{\raggedright\arraybackslash}p{0.39\textwidth}r@{}}
\toprule
Run & Elapsed & Waiting & Requests by reason & Interv. \\
\midrule
Pour-1 & 23:31 & 0:00 & none & 0 \\
Pour-2 & 12:31 & 1:24 & 1 clean-up & 0 \\
Pour-3 & 9:25 & 0:00 & none & 0 \\
Pour-C & 24:16 & 0:00 & none & 0 \\
Heat-1 & 35:39 & 0:00 & none & 0 \\
Heat-2 & 36:52 & 4:15 & 1 fault & 1 \\
Heat-C & 35:38 & 0:00 & none & 0 \\
Split-1 & 59:17 & 6:23 & 3 clean-up & 0 \\
Split-2 & 60:10 & 7:48 & 2 clean-up, 1 vessel & 0 \\
Split-3 & 34:27 & 0:00 & none & 0 \\
Split-4 & 31:34 & 3:20 & 1 permission & 0 \\
Split-5 & 24:03 & 0:00 & none & 0 \\
Split-C & 105:44 & 21:03 & 1 clean-up, 1 permission, 4 check & 0 \\
Calibrate-1 & 36:33 & 12:14 & 4 cable, 1 clean-up, 1 check & 0 \\
Calibrate-2 & 25:12 & 8:14 & 4 cable, 1 check & 0 \\
Calibrate-3 & 26:30 & 7:55 & 4 cable, 1 check & 0 \\
Calibrate-C & 48:26 & 8:43 & 3 check, 2 fault & 0 \\
Hydrogel-1 & 240:48 & 21:25 & 4 cable, 1 permission, 1 check & 0 \\
\bottomrule
\end{tabular}

\end{table}

\section{Materials and methods}
\label{si:methods}

\subsection{Bench and instruments}
\label{si:m-bench}

Every instrument was used as delivered. The arm, a Kinova Gen3 with a Robotiq 2F-85 parallel gripper, ran under ROS~2 Humble and MoveIt~2.5.9 on a Linux computer at the bench (the robot computer). A wrist camera, a RealSense D435 and a Logitech C920 viewed the bench, and the agent ran on a laptop connected to the robot computer over SSH. In every run an operator answered the agent's requests and labeled the criteria.

The balance reads to 1\,mg and streams its reading with a stability flag over RS-232, which the harness logged. The hot plate and stirrer (Fisherbrand Isotemp digital ceramic) has HEAT and STIR knobs, two seven-segment displays and no computer interface, so the arm turned the knobs while a camera read the displays. The potentiostat (Gamry Interface 1010E) was connected to a Windows~11 computer that ran the vendor's software and its Python toolkit, ToolkitPy, and the laptop reached this computer over SSH.

The vessels were a 50\,mL polypropylene cup, a 100\,mL glass beaker with a stir bar and plastic Petri dishes about 53\,mm across with loose lids. Two dishes, the molds, served as conductivity cells, each with two flat copper-foil electrodes on its bottom and cabled to the potentiostat as a two-electrode cell (Fig.~\ref{fig:si-molds}). One pair of molds served in \task{Calibrate} and a new pair in \task{Hydrogel}.

\subsection{Agent and session isolation}
\label{si:m-agent}

The agent has no memory of earlier runs beyond the skill store (Section~\ref{si:m-store}). Each run was a new non-interactive session of Claude Opus~5.5 at high reasoning effort, without saved sessions, MCP tool servers or permission prompts. The session received a fixed list of environment variables and ran in a mode that turns off user instructions, skills, hooks and memory. The prompt held the bench instructions, the task specification with the run's variant (Section~\ref{si:m-tasks}), the paths of the run's folders and its time cap.

The bench instructions put the task specifications, other runs' records and stores, memory and configuration folders and the experimenters' files off limits, a script checked every run's commands against these rules, and no run read another run's records or data files.

\subsection{Motion server and safety checks}
\label{si:m-motion}

The bench holds a glass beaker, a hot plate and liquids, and the agent writes its motion programs during a run. Every arm motion therefore passed through the motion server, one process with fixed safety checks that we provide and that the agent was told not to edit. The server, a ROS~2 process on the robot computer, reads requests as lines of JSON. A request stops the arm, operates the gripper or moves the arm. A move is of one of two kinds. In a Cartesian move the flange follows a straight line, and the MoveIt planner computes the joint motion for that line. In a joint-space move all joints move at proportional rates from their current angles to the target angles, and the server computes this line directly. The arm moves when the request asks for execution and the server's gate is open, and we opened the gate with an environment variable when we started the server.

The safety checks limit when a motion may start, how far and how fast the arm moves and where the tool may go. The server refused a motion when the gate was closed, when another motion was running or when its latest joint reading was older than 0.5\,s. It also refused a motion whose requested start lay more than 2$^\circ$ from the arm's joint angles, and one that left the joint limits or exceeded 30$^\circ$/s. A Cartesian move also had to be planned fully by MoveIt, stay within 0.20\,m and 20$^\circ$/s, move no joint more than 10$^\circ$ between waypoints and avoid the wrist singularity. For every joint-space move, and for a Cartesian move whose request gave limits, the server checked the path with forward kinematics (FK). The check keeps the tool center point above a floor height and outside the collision boxes that the agent sent with the request. The gripper closes at full force until it reaches a commanded position, and the pouring skill sets a speed of 8$^\circ$/s for moves that carry liquid.

The server accepts a stop at any time and halts the arm where it is. A stop sent over another connection stays in force until the operator releases it, and the arm also halts when the program that requested a motion disconnects.

\subsection{Execution harness}
\label{si:m-harness}

The harness has two parts (Fig.~2 of the main text). We provide the runner, the motion server, the timing wrapper (Section~\ref{si:m-time}), three skills and the bench instructions. The agent writes the skill store. During a run the agent calls the available tools and writes programs for the operations required by the task specification. The runner records the commands and their execution times, and the agent saves reusable programs and instructions in the skill store.

For the potentiostat, the Windows computer held the vendor's help pages and example scripts, including an example of electrochemical impedance spectroscopy (EIS) that measures impedance across frequencies. The agent used these resources to write the measurement program during \task{Calibrate}.

\subsection{The skill store}
\label{si:m-store}

The store holds a notes file, skills (written instructions for a kind of task), tools (programs, each described with how to call it and what it was tested on) and layout files (dated bench positions and calibrations). The agent was told to read the notes first and to write any capability it lacked into the store. The runner copied the store before and after every run (Fig.~2 of the main text), and each empty-store control wrote into a fresh store. As a supervisor checks an apprentice's notebook, we reviewed and corrected the store between runs, and the operator's answers entered it when the agent wrote them down. These corrections changed text in the notes, the layout files, two tool descriptions and one written skill, and none changed a program.

\subsection{Tasks and task specifications}
\label{si:m-tasks}

The last of the five tasks, a hydrogel workflow, combines the earlier four (Table~1 of the main text). Each task specification states the goal, starting state, parameters, criteria with their instruments, constraints and common rules (write a missing capability into the store, edit no harness file outside the store, record failures as failures), and the prompt carried one numbered variant.

In \task{Pour} the arm pours water from the cup into the glass beaker on the balance to a target mass. In \task{Heat} it heats the beaker on the hot plate through a 5\,min hold, which is assessed from the HEAT display, and reweighs the beaker. In \task{Split} it pours a target mass into each of two dishes on the balance pan without closing the gripper on a dish at full force. In \task{Calibrate} the agent writes a program that runs an impedance measurement and reads the result. With this program the agent determines each mold's cell constant $K$, which converts the mold's resistance $R$ into a conductivity. The reference is a 1000\,$\pm$\,10\,$\mu$S/cm sodium chloride standard that the operator filled into the molds. In \task{Hydrogel} the agent mixes PEDOT:PSS, a conductive polymer dispersed in water, with a precursor stock, cures the mixture in the molds on the hot plate and measures the gels (Section~\ref{si:m-hydrogel}).

The bench instructions told the agent to ask the operator for what lay outside the arm's work or was unexpected, such as refills, vessel exchanges, cable changes at the potentiostat and checks of the bench, and the operator carried out each request or declined it.

\subsection{Run protocol and controls}
\label{si:m-protocol}

Within a task, runs differed in the store, the target and the arrangement of the bench (Table~1 of the main text), and each was a new session with the same model and effort. Runs are named by task and order, with C for the empty-store control, and took place in the order of Table~2 of the main text. Each run was capped at 3\,h, and \task{Hydrogel} at 6\,h. The programs and skills that we provide stayed fixed within each task.

An empty-store control repeated a task's second variant from an empty store, after the task's runs with the store, with the same task specification, operator and bench. \task{Hydrogel} has no empty-store control due to the material cost.

\subsection{Time accounting}
\label{si:m-time}

A run's cost is the time the agent and the arm spend. A timing wrapper around every bench command recorded when the command started and ended and which category it belongs to. The categories are robot execution, instrument reading (the potentiostat included), procedure wait (heating up, the hold, stirring) and other tool execution. Waiting for the operator runs from each request to its answer and covers each logged intervention (Table~\ref{tab:si-help}). Each moment of a run counts in one category, the first that applies among waiting for the operator, procedure wait, robot execution, instrument reading and other tool execution. The remaining time is reasoning and programming, in which the agent reads, thinks, writes code and waits for the model. A procedure wait is a wait command that the agent ran through the wrapper. When the agent works during a hold, that time therefore counts as reasoning and programming, or in the category of the command that the agent runs. Commands that the agent sent to the robot computer outside the wrapper count as reasoning and programming.

We compare runs on agent-and-robot time, which is reasoning and programming plus robot execution, because the task specification fixes the procedure waits. Every run is timed over its whole session without the time waiting for the operator, and \task{Hydrogel} is timed to the end of its first round of measurements. Thinking tokens and API cost are those reported at the end of the session, at list prices, and cover the whole session. Bench commands are the shell commands that went through the timing wrapper or reached the robot computer. Section~\ref{si:times} gives every run's time in each category.

\subsection{Criteria, claims and labels}
\label{si:m-criteria}

Table~\ref{tab:si-criteria} gives the 19 criteria with their instruments and the operational definitions from the task specifications, and Table~\ref{tab:si-certify} lists what each instrument verifies. The \task{Heat} hold is assessed from the displayed plate temperature, and a spill means transferred water outside the receiver. The number of criteria met measures the task outcome.

For each experimental step, the task specifications required the agent to log a claim before reading the step's instrument through the wrapper. The claim states what the agent believed had happened and whether the step passed, failed or was uncertain. The agent then logged a check with the reading, the criterion and its verdict. The instruments, criteria and labels lay outside the agent's reach. After each run the operator labeled every criterion from the instrument records and the agent's checks.

\begin{table}[!htbp]
  \centering
  \caption{Success criteria with their instruments and the definition applied in labeling. Rows that name two dishes or molds are two criteria, which gives 19 in all. Targets are those of each run's variant (Table~1 of the main text).}
  \label{tab:si-criteria}
  \footnotesize
  \begin{tabular}{@{}>{\raggedright\arraybackslash}p{0.10\textwidth}>{\raggedright\arraybackslash}p{0.13\textwidth}>{\raggedright\arraybackslash}p{0.20\textwidth}>{\raggedright\arraybackslash}p{0.48\textwidth}@{}}
    \toprule
    Task & Criterion & Instrument & Definition \\
    \midrule
    \task{Pour} & mass & balance & Stable reading after the pour minus before, within the target $\pm$\,1.0\,g \\
     & no spill & camera frames and the operator & No transferred water outside the receiving beaker \\
    \midrule
    \task{Heat} & setpoint & HEAT display & Setpoint shown when the knob stops \\
     & hold & HEAT display & Displayed plate temperature within $\pm$\,1\,$^\circ$C of the setpoint for at least 60\,s of the 5\,min hold \\
     & off & HEAT display & OFF shown when HEAT is turned off \\
     & reweigh & balance & At most 1\,g lost between leaving the pan and returning to it \\
    \midrule
    \task{Split} & dish 1, dish 2 & balance & Water in the dish within the target $\pm$\,0.5\,g \\
     & no spill & camera frames and the operator & No transferred water outside the dishes \\
    \midrule
    \task{Calibrate} & program & data file with the run's time stamp & The agent's program starts a measurement and reads its result back, with no click in the vendor's software \\
     & cell constants & potentiostat data & A cell constant for each mold from the 1000\,$\mu$S/cm standard, with its readings \\
    \midrule
    \task{Hydrogel} & PEDOT:PSS & balance & 16.0\,$\pm$\,0.5\,g added \\
     & stir & STIR display & 800\,$\pm$\,10\,rpm for at least 2\,min, then 0 \\
     & mold 1, mold 2 & balance & 6.0\,$\pm$\,1.0\,g of mixture poured into the mold on the pan \\
     & lids & camera frames & Both molds on the hot plate during the heating, lids on and not fully closed \\
     & heat & HEAT display & HEAT display at 60\,$\pm$\,1\,$^\circ$C during the 2\,h hold, then OFF \\
     & gelled & camera frames and the operator & Both molds gelled \\
     & conductivity & potentiostat data and balance & A conductivity for each mold from $R$ and $K$ at the weighed mass, with its readings, and the difference between the molds \\
    \bottomrule
  \end{tabular}
\end{table}

\begin{table}[!htbp]
  \centering
  \caption{What each instrument verifies.}
  \label{tab:si-certify}
  \small
  \begin{tabular}{@{}>{\raggedright\arraybackslash}p{0.27\textwidth}>{\raggedright\arraybackslash}p{0.67\textwidth}@{}}
    \toprule
    Instrument & Verifies \\
    \midrule
    Balance & mass poured (\task{Pour}, \task{Split}, \task{Hydrogel}), mass lost on heating (\task{Heat}) and mass of each gel after curing (\task{Hydrogel}) \\
    Hot-plate displays, read in camera frames & setpoint, hold and OFF (\task{Heat}, \task{Hydrogel}), and stirring speed and its return to 0 (\task{Hydrogel}) \\
    Cameras & no water outside the receiving vessels (\task{Pour}, \task{Split}), and lids on during the heating and gels formed (\task{Hydrogel}) \\
    Potentiostat & a data file written and read back by the agent's program, and the impedance of the standard in the molds (\task{Calibrate}) and of the gels (\task{Hydrogel}) \\
    \bottomrule
  \end{tabular}
\end{table}

\subsection{Hydrogel, molds and conductivity}
\label{si:m-hydrogel}

The recipe came from a collaborator. The operator put 0.90\,g of stock (acrylic acid, water, ammonium persulfate as initiator and \textit{N,N}$'$-methylenebisacrylamide as cross-linker, 100:100:1:1 by mass) and a stir bar into the glass beaker and about 18\,g of PEDOT:PSS dispersion into the cup. The task specification's amounts are in Table~1 of the main text. For comparison, we prepared one gel by hand from the same recipe, poured 5.96\,g of the mixture into the second mold and cured it for 2\,h on the hot plate at 60\,$^\circ$C.

The agent's program measured each gel by EIS as a two-electrode cell, with two sweeps from 100\,kHz to 100\,Hz (16 frequencies, 0\,V DC, 10\,mV rms). The solution resistance $R$ is the real part of the impedance at 63\,kHz, where the sweeps of the standard level off at the solution resistance with a phase between $-1.0^\circ$ and $-0.3^\circ$. For the gels the phase there was $-1.7^\circ$ and $-2.4^\circ$. The conductivity is $\kappa = K/R$, with $K$ taken at the gel's weighed mass and the standard's conductivity taken at 25\,$^\circ$C.

For the gels we calibrated the molds with standard before the \task{Hydrogel} run (Fig.~\ref{fig:si-cells}). With each mold on the balance pan, the standard was added in nominal 1\,g steps from 4 to 9\,g and stirred after each step, as pouring stirs the gel mixture. Two sweeps from 1\,MHz to 1\,Hz followed about a minute later. Each $K = R\,\kappa_{\mathrm{std}}$ uses the mean $R$ of the two sweeps, with $\kappa_{\mathrm{std}}$ the standard's 1000\,$\mu$S/cm.

\section{Comparison with related systems}
\label{si:related}

Table~\ref{tab:si-related} compares LabApprentice with systems in which agents program robots or reuse experience.

{\scriptsize
\setlength{\tabcolsep}{4pt}
\begin{longtable}{@{}>{\raggedright\arraybackslash}p{0.125\textwidth}>{\raggedright\arraybackslash}p{0.15\textwidth}>{\raggedright\arraybackslash}p{0.17\textwidth}>{\raggedright\arraybackslash}p{0.155\textwidth}>{\raggedright\arraybackslash}p{0.165\textwidth}>{\raggedright\arraybackslash}p{0.14\textwidth}@{}}
  \caption{LabApprentice and related systems. The column Experience comparison names what each system was compared against, such as an empty store, an initial stack of programs or a learned store of text. Not reported means that the paper does not describe such a comparison.}\label{tab:si-related}\\
    \toprule
    System & Setting & Who writes the skills & Reuse across fresh sessions & Experience comparison & Instrument operation \\
    \midrule
    \endfirsthead
    \toprule
    System & Setting & Who writes the skills & Reuse across fresh sessions & Experience comparison & Instrument operation \\
    \midrule
    \endhead
    \bottomrule
    \endfoot
    AGP~\cite{agp2026} & real arms: assembly, blocks, dice, throwing, towel folding & agent (programs, measurements, lessons) & yes, same task, fresh context per run, also handed to a weaker model & weaker model, 5 runs with and 5 without, same-model repeats noted as needed & none in its tasks \\
    \addlinespace[3pt]
    WetRobo~\cite{wetrobo2026} & real arm in two biology laboratories: Petri lid, bottle cap, incubator door & agent (programs) and people (skill file, base code) & yes, programs kept in the repository and reused by later tasks & not reported & incubator door, opened by its handle \\
    \addlinespace[3pt]
    LabEvolver~\cite{labevolver2026} & real arm on a wetlab bench (pouring, pH and pH--conductivity adjustment) and ALFWorld & people (skills, safety rules) and agent (experience entries, skill parameters) & yes, retrieved before each trial, also across tasks & yes, no-experience and inner-loop-only runs & balance and water-quality meter, interface not reported \\
    \addlinespace[3pt]
    ASPIRE~\cite{aspire2026} & simulation benchmarks, real bimanual YAM manipulation and Rubik's Cube & agent (validated repair guidance, code sketches and online text skills) & source skills guide new tasks and embodiments, cube episodes share one session & empty-library simulation, physical transfer with and without the selected simulation skill & no laboratory instruments in its real-robot tasks \\
    \addlinespace[3pt]
    Recursia~\cite{recursia2026} & simulation benchmarks, three compositional tasks on a real Franka FR3 & agent (shared tools, skill contracts, handoffs and verifiers), retained after validation & frozen source-trained stack reused on new target compositions & initial stack in simulation, physical comparator retains a learned skill document & no laboratory instruments in its physical tasks, success manually assessed \\
    \addlinespace[3pt]
    ELLMER~\cite{ellmer2025} & Kinova Gen3 arm: coffee making, plate decoration & people (curated code examples), GPT-4 adapts them per task & curated examples retrieved per query, updating from new experience left to future work & not reported (baseline kept the examples in its context) & none in its tasks \\
    \addlinespace[3pt]
    Voyager~\cite{voyager2023} & Minecraft (video game) & agent (programs on primitives written by the authors) & yes, library reused in a new world on unseen tasks & yes, without the skill library & not applicable \\
    \addlinespace[3pt]
    Coscientist~\cite{boiko2023} & liquid handler, plate reader, cloud laboratory & agent (code for each task, from API documentation) & not reported & not reported (optimization run with and without 10 prior yields) & devices reached through APIs \\
    \addlinespace[3pt]
    ORGANA~\cite{organa2024} & Franka arm: solubility, recrystallization, pH, electrochemistry & people (skills, planning domain) and LLM (experiment plans) & within one series of experiments, across sessions not reported & not reported & hot plate over PyVISA, scales and potentiostat over USB \\
    \addlinespace[3pt]
    EmbodiRSI~\cite{embodirsi2026} & three real tabletop scenes copied in simulation, real dual-arm robot & none: agents generate simulated data on which a policy is fine-tuned & policy and data across rounds within a scene & adaptive against static data collection & none \\
    \addlinespace[3pt]
    EmbodiedEvo \cite{embodiedevo2026} & simulation benchmark & agent (code skills and text lessons, one scene per task) & yes, frozen store given to fresh agents in unseen scenes & yes, with and without the store, same model & none \\
    \addlinespace[3pt]
    EmbodiedRSI \cite{embodiedrsi2026} & simulation benchmarks, frozen harness then run on a real arm & agent (code and skills evolved in simulation) & within one evolution run, simulation to real arm & not reported with an empty memory & none \\
    \addlinespace[3pt]
    CaP-X~\cite{capx2026} & simulation benchmarks, real Franka arm and AgiBot G1 & agent (a program per task), nine functions synthesized once from pooled simulated rollouts & library fixed after a single synthesis pass & skill-library ablation in simulation, shown in a figure & none \\
    \addlinespace[3pt]
    Guava~\cite{guava2026} & simulation, real Franka arm for evaluation & none: tools of the harness written by the authors, a compact model fine-tuned on simulated trajectories & none, history within one episode & base against fine-tuned model & none \\
    \addlinespace[3pt]
    ARCHITECT~\cite{architect2026} & real Franka arm, eight manipulation tasks with articulated and deformable objects & agent (programs), skills distilled from a person's corrections & yes, library carried from one task to the next & yes, empty against populated library (human queries per trial) & none \\
    \addlinespace[3pt]
    LEXP~\cite{lexp2026} & 12 real manipulation tasks & agent (lessons extracted from execution evidence), model weights fixed & yes, lessons reused across episodes & yes, without retained experience on three tasks & none \\
    \addlinespace[3pt]
    LabApprentice & wetlab bench & agent (code, skills, layouts) and people (base programs and skills) & yes, across tasks & yes, every repeated task & hot plate (knob, display), balance over its serial stream, potentiostat through its vendor toolkit \\
\end{longtable}
}

\section{Time of every run by category}
\label{si:times}

\begin{figure}[!htb]
  \centering
  \maybefig{width=0.64\textwidth}{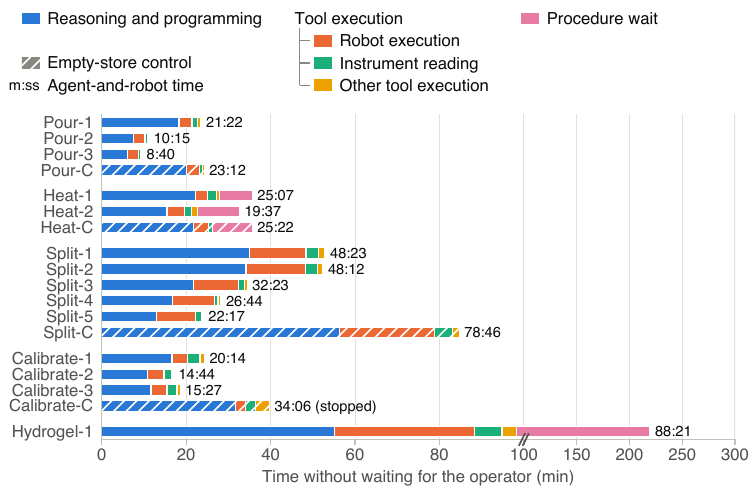}
  \caption{Time of every run without waiting for the operator, in run order per task. Each bar is split into reasoning and programming, tool execution (robot execution, instrument reading, other tool execution) and procedure wait (heating up, the hold, stirring). Empty-store controls are hatched, and the label at each bar's end gives the run's agent-and-robot time. The \task{Calibrate}-C run is marked stopped, and its bar ends at the stop. Beyond 100\,min the time axis is compressed four-fold. Table~\ref{tab:si-time} gives the values.}
  \label{fig:si-times}
\end{figure}

Figure~\ref{fig:si-times} shows the time of every run in each category of Section~\ref{si:m-time}. Table~\ref{tab:si-time} gives the values behind the figure, which include the times quoted in Sections~4.3 and~4.4 of the main text.

\begin{table}[!htb]
  \centering
  \caption{Time of every run in each category, in min:s, without waiting for the operator. R\&P is reasoning and programming, Robot is robot execution, and Agent+robot is agent-and-robot time, the sum of the two (Table~2 of the main text). Instr.\ is instrument reading, Other is other tool execution, Wait is procedure wait (heating up, the hold, stirring) and Total is the whole timed period. Each value is rounded to the second, so the parts of a row can differ from its total by a second. \task{Hydrogel} is timed to the end of its first round of measurements, and \task{Calibrate}-C to the stop.}
  \label{tab:si-time}
  \scriptsize
  \begin{tabular}{@{}lrrrrrrr@{}}
\toprule
Run & R\&P & Robot & Agent+robot & Instr. & Other & Wait & Total \\
\midrule
Pour-1 & 18:17 & 3:05 & 21:22 & 1:17 & 0:52 & 0:00 & 23:31 \\
Pour-2 & 7:34 & 2:41 & 10:15 & 0:42 & 0:09 & 0:00 & 11:07 \\
Pour-3 & 6:06 & 2:34 & 8:40 & 0:34 & 0:11 & 0:00 & 9:25 \\
Pour-C & 20:06 & 3:06 & 23:12 & 0:39 & 0:25 & 0:00 & 24:16 \\
\addlinespace
Heat-1 & 22:13 & 2:55 & 25:07 & 2:03 & 0:46 & 7:42 & 35:39 \\
Heat-2 & 15:29 & 4:08 & 19:37 & 1:42 & 1:18 & 9:59 & 32:37 \\
Heat-C & 21:49 & 3:32 & 25:22 & 0:52 & 0:06 & 9:18 & 35:38 \\
\addlinespace
Split-1 & 34:59 & 13:24 & 48:23 & 2:53 & 1:38 & 0:00 & 52:53 \\
Split-2 & 34:10 & 14:02 & 48:12 & 2:56 & 1:13 & 0:00 & 52:22 \\
Split-3 & 21:40 & 10:44 & 32:23 & 1:23 & 0:40 & 0:00 & 34:27 \\
Split-4 & 16:51 & 9:53 & 26:44 & 0:49 & 0:41 & 0:00 & 28:14 \\
Split-5 & 13:03 & 9:14 & 22:17 & 1:27 & 0:19 & 0:00 & 24:03 \\
Split-C & 56:19 & 22:28 & 78:46 & 4:22 & 1:32 & 0:00 & 84:41 \\
\addlinespace
Calibrate-1 & 16:37 & 3:37 & 20:14 & 3:03 & 1:02 & 0:00 & 24:19 \\
Calibrate-2 & 10:46 & 3:59 & 14:44 & 1:55 & 0:19 & 0:00 & 16:58 \\
Calibrate-3 & 11:42 & 3:45 & 15:27 & 2:23 & 0:46 & 0:00 & 18:35 \\
Calibrate-C & 31:39 & 2:27 & 34:06 & 2:23 & 3:15 & 0:00 & 39:43 \\
\addlinespace
Hydrogel-1 & 55:09 & 33:11 & 88:21 & 6:27 & 3:23 & 121:13 & 219:24 \\
\bottomrule
\end{tabular}

\end{table}

\section{Operation types exercised by the tasks}
\label{si:coverage}

Table~\ref{tab:si-coverage} gives, for three annotated corpora of laboratory procedures, the number of annotated operations of each type and the tasks that exercise the type.

\begin{table}[!htb]
  \centering
  \caption{Operation types in three annotated corpora and the tasks that exercise them. Counts are those of Table~1 of ref.~\cite{tamari2021xwlp} (279 wetlab biochemistry protocols), of Table~1 of ref.~\cite{wang2022ulsa} (535 inorganic synthesis paragraphs) and of Table~5 of ref.~\cite{vaucher2020actions} (352 sentences of organic synthesis procedures from patents, where the count of a type is the number of its actions in the ground truth). The types that the tasks exercise account for 2481 of 3915 operations (63\%), 3364 of 5547 actions (61\%) and 514 of 939 actions (55\%).}
  \label{tab:si-coverage}
  \scriptsize
  \setlength{\tabcolsep}{4pt}
  \begin{minipage}[t]{0.51\textwidth}
    \centering
    \begin{tabular}[t]{@{}lr>{\raggedright\arraybackslash}p{0.42\textwidth}@{}}
      \toprule
      \multicolumn{3}{@{}l}{Biochemistry protocols~\cite{tamari2021xwlp}} \\
      Type & Count & Tasks \\
      \midrule
      Transfer & 1301 & all five \\
      Temperature Treatment & 503 & \task{Heat}, \task{Hydrogel} \\
      General & 469 & \\
      Mix & 346 & \task{Hydrogel} \\
      Spin & 282 & \\
      Create & 178 & \\
      Destroy & 170 & \\
      Remove & 168 & \\
      Measure & 149 & all five \\
      Wash & 146 & \\
      Time & 114 & \task{Heat}, \task{Hydrogel} \\
      Seal & 68 & \task{Hydrogel} \\
      Convert & 21 & \\
      \midrule
      Total & 3915 & \\
      \bottomrule
    \end{tabular}

    \vspace{10pt}
    \begin{tabular}[t]{@{}lr>{\raggedright\arraybackslash}p{0.42\textwidth}@{}}
      \toprule
      \multicolumn{3}{@{}l}{Inorganic synthesis~\cite{wang2022ulsa}} \\
      Type & Count & Tasks \\
      \midrule
      Starting & 619 & \\
      Mixing & 1853 & \task{Hydrogel} \\
      Purification & 1080 & \\
      Heating & 973 & \task{Heat}, \task{Hydrogel} \\
      Cooling & 259 & \\
      Shaping & 225 & \\
      Reaction & 232 & \task{Heat}, \task{Hydrogel} \\
      Non-Altering & 306 & \task{Heat}, \task{Split}, \task{Calibrate}, \task{Hydrogel} \\
      \midrule
      Total & 5547 & \\
      \bottomrule
    \end{tabular}
  \end{minipage}\hfill
  \begin{minipage}[t]{0.45\textwidth}
    \centering
    \begin{tabular}[t]{@{}lr>{\raggedright\arraybackslash}p{0.40\textwidth}@{}}
      \toprule
      \multicolumn{3}{@{}l}{Organic synthesis~\cite{vaucher2020actions}} \\
      Type & Count & Tasks \\
      \midrule
      Add & 255 & \task{Pour}, \task{Split}, \task{Hydrogel} \\
      Stir & 118 & \task{Hydrogel} \\
      MakeSolution & 62 & \task{Hydrogel} \\
      SetTemperature & 60 & \task{Heat}, \task{Hydrogel} \\
      Concentrate & 54 & \\
      Wash & 45 & \\
      PH & 43 & \\
      CollectLayer & 37 & \\
      Yield & 37 & \\
      Extract & 34 & \\
      Filter & 34 & \\
      NoAction & 25 & \\
      Purify & 24 & \\
      DrySolution & 22 & \\
      Wait & 19 & \task{Heat}, \task{Hydrogel} \\
      FollowOtherProcedure & 15 & \\
      DrySolid & 12 & \\
      InvalidAction & 11 & \\
      Quench & 8 & \\
      Reflux & 7 & \\
      Partition & 5 & \\
      PhaseSeparation & 4 & \\
      Triturate & 3 & \\
      OtherLanguage & 2 & \\
      Recrystallize & 2 & \\
      Degas & 1 & \\
      \midrule
      Total & 939 & \\
      \bottomrule
    \end{tabular}
  \end{minipage}
\end{table}

\renewcommand{\bibfont}{\small}
\bibliographystyle{ieeenat_fullname}
\bibliography{references}